\documentclass[runningheads]{llncs}
\usepackage{accv}

\usepackage{accvabbrv}

\usepackage{graphicx}
\usepackage{booktabs}
\usepackage{multirow}
\usepackage{placeins}
\usepackage{pgfplots}
\pgfplotsset{compat=1.16}

\usepackage[accsupp]{axessibility}  

\usepackage[breaklinks,colorlinks,citecolor=accvblue]{hyperref}

\usepackage{orcidlink}

\begin{document}

\title{StimFlow: Synthesizing Brain Activity from Images via Stimulus-conditioned Flow Matching}

\titlerunning{StimFlow: Stimulus-conditioned Flow Matching for fMRI Synthesis}

\author{Bui Van Hung\inst{1} \and Dang Thanh Vu\inst{2} \and Hoang-Son Vo\inst{1} \and Mai Duc Minh Huy\inst{3} \and Mai Tien Dung\inst{4} \and Soo-Hyung Kim\inst{1}}

\authorrunning{B.~V.~Hung et al.}

\institute{Chonnam National University, Gwangju, South Korea \and AISEED, Inc., South Korea \and Ho Chi Minh University of Science, Ho Chi Minh City, Vietnam \and MIREA -- Russian Technological University, Moscow, Russia}

\maketitle

\begin{abstract}
  The inherent variability of fMRI responses across trials encourages a
  generative view of the stimulus-to-response mapping. Prior generative encoders
  often operate in compressed latent spaces and condition predominantly on
  semantic features, which provide limited low-level cues for early visual
  cortex. We propose \textbf{StimFlow}, which synthesizes fMRI directly in
  \textbf{native voxel space} using \textbf{conditional flow matching} with a
  linear path. This design aligns training and inference distributions, enabling
  competitive performance with a single sample and reducing reliance on
  averaging. To match the visual hierarchy, StimFlow combines structural
  \textbf{DINOv2} features with semantic \textbf{CLIP} features, which prove
  complementary along the cortical processing stream. With a single sample,
  StimFlow achieves a per-voxel Pearson correlation of $\mathbf{0.371}$,
  matching baselines that average five draws. Generation in voxel space allows
  direct ROI analysis and concept localization without a decoder. Furthermore,
  it recovers over half of full-data performance from just one hour of scans,
  supporting efficient cross-subject adaptation.
\end{abstract}

\section{Introduction}
\label{sec:intro}
Understanding how the human brain encodes natural visual stimuli is a central goal in
computational neuroscience and brain--computer interfaces~(BCI)~\cite{naselaris2011,Mitchell2008PredictingHB,Haxby2001DistributedAO}.
Owing to physiological noise, cognitive-state fluctuations, and hemodynamic variability,
the stimulus-to-response mapping is inherently stochastic~\cite{nsd}; even the more modest
goal of approximating the \emph{conditional mean} benefits from a generative formulation
whose training and inference distributions are aligned, rather than from a deterministic
regressor or a diffusion prior whose higher-curvature paths necessitate multi-sample
averaging for stability.

Recent generative encoders pursue this goal but share two limitations. First, they
synthesize in a \emph{compressed, CLIP-aligned latent space} and are evaluated mainly through a
decoded image; because the signal is recovered through an autoencoder rather than generated in
voxel space, fidelity at the voxel and ROI level remains largely unexamined. Second, they
condition \emph{exclusively} on language-supervised CLIP features~\cite{clip}, which favour
high-level semantics and retain little of the low-level structure (edges, textures, spatial
frequency) driving early visual areas (V1--V3), so a single semantic stream is poorly matched
to the full functional hierarchy of the visual cortex.

The two representative encoders also differ in how they reach the data distribution:
MindSimulator~\cite{mindsimulator} places a diffusion prior over fMRI autoencoder latents
and relies on multi-trial averaging and post-selection to obtain stable predictions,
whereas SynBrain~\cite{synbrain} uses a one-step mapper that is stable but acts as a
deterministic predictor. We seek a formulation whose training and inference paths are
aligned by construction, so that a single sample ($T{=}1$) already yields a faithful
estimate of the conditional mean without post-hoc averaging.

We propose \textbf{StimFlow}, which addresses both axes. StimFlow casts fMRI synthesis as
Conditional Flow Matching~(CFM)~\cite{lipman2023,rectifiedflow} along a conditional
linear path with independent coupling, generating directly in \emph{native voxel space} with a 1D Diffusion
Transformer~(DiT-1D)~\cite{dit}; because the training interpolation starts exactly at the
Gaussian distribution inference begins from, training and inference are aligned by
construction. To match the cortical hierarchy, it pairs a structural stream
(DINOv2~\cite{dinov2}) with a semantic stream (CLIP), which we find functionally
complementary: structural features lead in early-to-mid visual areas, semantic
features in higher ventral and lateral streams. A shared trunk with lightweight per-subject
adapters further enables few-shot adaptation to novel subjects.

On the Natural Scenes Dataset~\cite{nsd}, a \emph{single} StimFlow sample reaches
a per-voxel Pearson correlation of $0.371$, matching or exceeding baselines that average
five draws at the voxel level.
More importantly, because generation stays in voxel space, fidelity can be read directly
per ROI without a separate decoder, and synthesized concept-selectivity maps align closely
with the ground truth ($r=0.94$, top-$10\%$ Dice $=0.81$), supporting data-driven cortical analysis.

In summary, our contributions are:
\begin{itemize}
  \item We formulate fMRI synthesis as \textbf{conditional flow matching in native voxel
    space}, enabling direct per-ROI readout and concept localization without a decoder;
    a single sample matches or exceeds multi-sample baselines on NSD~\cite{nsd}
    at both the voxel and semantic level.
  \item We show that \textbf{CLIP-only conditioning is suboptimal}: a structural
    stream (DINOv2) matches or exceeds CLIP alone, and the two are functionally
    complementary along the cortical hierarchy, motivating dual-stream conditioning.
  \item We \textbf{adapt the DiT architecture to 1D voxel sequences and equip it with
    per-subject adapters}, supporting few-shot adaptation to novel subjects.
\end{itemize}

\section{Related Work}
\label{sec:related}

\paragraph{Brain visual encoding.}
Visual encoding models predict voxel-level fMRI responses from images and have long been used
to probe the functional organization of visual cortex~\cite{naselaris2011}. Early work fit
explicit feature spaces (Gabor-wavelet, population receptive-field models) by linear
regression~\cite{kay2008,Haxby2001DistributedAO,Mitchell2008PredictingHB}; the deep-learning
era replaced these with hierarchical CNN activations predicting ventral-stream
responses~\cite{yamins2014,guclu2015,algonauts2023}, and recent encoders use self-supervised
(DINOv2~\cite{dinov2}) or language-supervised (CLIP~\cite{clip}) features. Yet these encoders
remain deterministic, mapping each image to a single response and thus unable to express the
trial-to-trial variability intrinsic to fMRI~\cite{nsd}.

\paragraph{Brain decoding and reconstruction.}
The inverse problem (reconstructing the seen image from fMRI) has progressed from early movie
reconstruction~\cite{Nishimoto2011} to diffusion-prior
methods~\cite{takagi2023,mindeye,mindeye2} that map fMRI into CLIP space and condition an image
generator, with MindEye2 enabling strong low-data cross-subject transfer. Decoding is
complementary to our goal of synthesizing responses from stimuli; we reuse a frozen
MindEye2~\cite{mindeye2} only to decode synthesized responses and check their semantic content
(\cref{sec:exp}).

\paragraph{Generative fMRI synthesis.}
Closest to us, a small line of work models the stimulus-to-response mapping as a conditional
distribution. MindSimulator~\cite{mindsimulator} learns a diffusion prior over fMRI-autoencoder
latents and SynBrain~\cite{synbrain} a one-step semantic-to-neural VAE mapper; both operate in
a compressed, CLIP-conditioned latent space, sharing the two limitations we target (no direct
per-ROI readout, a semantic-only bias) and a stability trade-off: the diffusion prior
requires multi-sample averaging for stable results, while the one-step mapper is stable but
deterministic. StimFlow instead synthesizes in native voxel space with a complementary DINOv2
stream and sidesteps the trade-off via flow matching; we compare against both in \cref{sec:exp}.

\paragraph{Visual representations and the semantic bias of CLIP.}
Which feature space best matches visual cortex remains open. CLIP's language supervision makes
it predictive of high-level, category-selective regions (FFA~\cite{kanwisher1997ffa},
PPA~\cite{epstein1998ppa}) but discards low-level structure driving early areas V1--V3, whereas
self-supervised transformers such as DINOv2~\cite{dinov2,dosovitskiy2021image} retain fine
spatial structure; supervised and self-supervised models are known to correlate differently
with cortical hierarchies~\cite{Khaligh-Razavi2014,chen2020simclr,He_2022_CVPR}. This motivates
the multi-stream conditioning we analyse per-ROI in \cref{sec:exp-abl}, where the two
representations prove functionally complementary rather than interchangeable.

\paragraph{Flow matching and diffusion transformers.}
Flow matching~\cite{lipman2023} regresses a velocity field onto a probability path; a conditional linear path with independent noise--data coupling yields straight conditional trajectories whose lower curvature allows stable integration with fewer ODE steps than typical diffusion samplers~\cite{rectifiedflow,song2021scorebased}. The Diffusion Transformer~(DiT) replaces the U-Net with a transformer conditioned via AdaLN~\cite{dit,esser2024sd3}. Within neuroimaging, flow and rectified-flow models have so far been applied to medical image synthesis~\cite{fm_medsyn2025} and to brain-connectivity matrices on Riemannian manifolds~\cite{diffeocfm2025}, but to our knowledge not yet to stimulus-conditioned, voxel-space fMRI encoding. StimFlow adapts DiT to 1D fMRI sequences under conditional flow matching to model visual-to-neural mapping in native space, and is among the first to bring CFM to this setting.

\section{Method}
\label{sec:method}

Two properties of fMRI data motivate our design.
First, fMRI voxels are indexed by anatomical location rather than a regular 2D pixel grid,
so flattening them into a 1D token sequence imposes far less artificial structure, making a
1D Transformer the appropriate backbone.
Second, we want the training and inference distributions to match exactly: conditional
flow matching achieves this because the interpolation starts at the same Gaussian
distribution inference begins from, and the same ODE is used at both stages. In practice
this \emph{train--inference alignment} yields lower-curvature trajectories than a
VP-SDE diffusion path~\cite{rectifiedflow}, allowing stable integration with fewer
steps.

\subsection{Problem Formulation}
\label{sec:method-prob}

We consider the Natural Scenes Dataset~(NSD)~\cite{nsd}, which provides paired data
$\{(\mathbf{x}_i, \mathbf{v}_i)\}_{i=1}^{N}$ where $\mathbf{x}_i$ denotes a natural
image and $\mathbf{v}_i \in \mathbb{R}^{n_s}$ is the corresponding z-scored fMRI
response vector for subject~$s$, comprising all voxels in the \texttt{nsdgeneral} ROI
($n_s$ ranges from roughly 12\,000 to 16\,000 across subjects).
Because identical stimuli elicit variable responses across trials, this mapping is
inherently stochastic; our goal is to learn the conditional distribution
$p(\mathbf{v}\mid\mathbf{x})$ directly in \emph{native voxel space}, synthesizing
high-fidelity fMRI responses for any image.

\subsection{Native-Space Conditional Flow Matching}
\label{sec:method-flow}

We cast fMRI synthesis as a conditional flow matching~(CFM)
problem~\cite{lipman2023,rectifiedflow}.
Let $\mathbf{x}_0 \sim \mathcal{N}(\mathbf{0}, \mathbf{I})$ denote a pure noise sample
in $\mathbb{R}^{n_s}$ and $\mathbf{v}$ the measured fMRI target.
We adopt the \emph{conditional linear} probability path (independent coupling~\cite{lipman2023})
\begin{equation}
  \mathbf{x}_t = (1 - t)\,\mathbf{x}_0 + t\,\mathbf{v}, \qquad t \in [0, 1],
  \label{eq:xt}
\end{equation}
whose conditional velocity field is simply
$\mathbf{u}_t = \mathbf{v} - \mathbf{x}_0$.
A neural network $f_\theta$ is trained to predict this velocity, conditioned on the
interpolated state~$\mathbf{x}_t$, the time~$t$, and a stimulus
representation~$c(\mathbf{x})$ (described in \cref{sec:method-cond}):
\begin{equation}
  \mathcal{L}_{\text{FM}}
    = \mathbb{E}_{t,\,\mathbf{x}_0,\,(\mathbf{x},\mathbf{v})}\;
      \frac{1}{|\Omega|}\sum_{j\in\Omega} w_j\,
      \big( f_\theta(\mathbf{x}_t,\, t,\, c(\mathbf{x}))_j
        - (\mathbf{v} - \mathbf{x}_0)_j \big)^2,
  \label{eq:fm}
\end{equation}
where $\Omega$ is the set of real (non-padded) voxels and $w_j$ is a per-voxel
weight set to voxel $j$'s noise-ceiling reliability, normalised to unit mean
over $\Omega$, so that the loss emphasises reliably measured voxels and ignores
padding.
The reliability is estimated \emph{exclusively from training-set} repeated-stimulus
trials (split-half correlation~\cite{nsd}); no test-set data enter the weight
computation, so there is no information leakage from evaluation stimuli. The training time $t$ is drawn from a logit-normal
schedule~\cite{esser2024sd3} that concentrates samples away from the
$t\!\to\!0,1$ endpoints; its exact parameters are given in \cref{sec:supp-loss}.

The linear path choice is deliberate. The conditional interpolant in \cref{eq:xt} is
straight by construction; while averaging over the conditional posterior bends the learned
\emph{marginal} trajectory, a linear coupling induces lower curvature than a VP-SDE diffusion
path~\cite{rectifiedflow}, letting inference integrate the ODE stably with a modest number of
Euler steps ($N_\text{step}{=}50$, an order of magnitude below typical score-based samplers).
The whole process runs in native voxel space, so no autoencoder is required and per-ROI
analysis can be performed directly on the output.

\subsection{Stimulus Conditioning}
\label{sec:method-cond}

The visual cortex spans a hierarchy from oriented edges in V1 to semantic categories in areas
such as the fusiform face area~(FFA). A single CLIP embedding, as used by prior generative
encoders, is biased toward the semantic end and underserves early areas, so StimFlow
conditions on two complementary streams.

\paragraph{Global semantic context (CLIP).}
A pooled $1280$-d embedding from a frozen SDXL-CLIP encoder, projected by a two-layer MLP and
summed with the timestep embedding, forms the global signal
$\mathbf{c}=\mathbf{e}_t+\mathbf{e}_{\text{CLIP}}$ that modulates every transformer block
through Adaptive Layer Normalization~(AdaLN). Because CLIP provides a single scene-level
summary, AdaLN, which uniformly scales and shifts each layer's activations, is a natural fit,
setting the overall semantic context without attending to specific voxel positions.

\paragraph{Structural context (DINOv2).}
We extract multi-layer grid features from a frozen DINOv2-ViT-G/14
backbone~\cite{dinov2,dosovitskiy2021image}. Unlike CLIP, these self-supervised features
retain spatial structure and fine-grained properties (texture, contour, part--whole
relations), spanning low-level structure to emerging semantics across layers. After linear
projection and RMSNorm, the token sequence enters each block via multi-head cross-attention,
so each voxel patch attends to the most relevant spatial features. Cross-attention is preferred
over concatenation or global pooling because the feature map is spatially structured and
different cortical regions need different subsets of it, which a single global projection
would collapse.

\begin{figure}[tb]
  \centering
  \includegraphics[width=\linewidth]{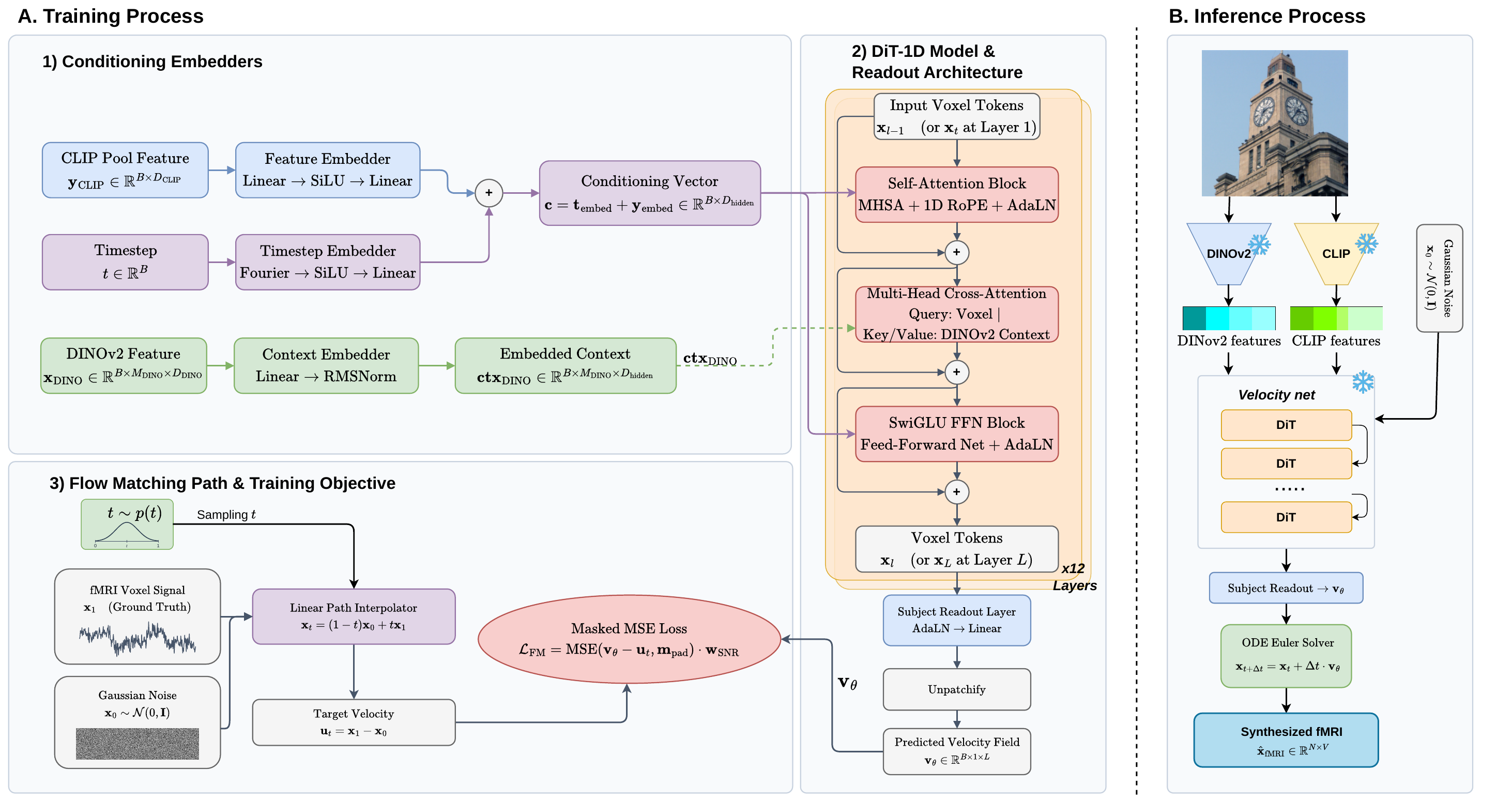}
  \caption{\textbf{Overview of the StimFlow training and inference framework.} (a)~\emph{Training Process}: Conditioning vectors $\mathbf{c}$ (combining CLIP semantics and timestep information) are derived from the input image, while structural context from DINOv2 is cross-attended. Voxel tokens are patchified and processed through $12$ DiT-1D layers to regress the conditional velocity field $\mathbf{u}_t$ directly in native voxel space using a masked MSE loss. (b)~\emph{Inference Process}: Starting from Gaussian noise $\mathbf{x}_0$, the velocity field is integrated over time using an ODE Euler solver to synthesize the final fMRI response.}
  \label{fig:fm}
\end{figure}

\subsection{DiT-1D Velocity Network}
\label{sec:method-arch}

The velocity field~$f_\theta$ is parameterised by a \emph{1D Diffusion
Transformer}~(DiT-1D), adapted from the DiT architecture~\cite{dit} to operate on
sequential rather than spatial data.
The design proceeds in three stages.

\paragraph{Patchification.}
The z-scored voxel vector~$\mathbf{v} \in \mathbb{R}^{L}$ ($L = 16384$ after zero-padding) is
treated as a single-channel 1D signal and tokenised into non-overlapping patches via a strided
Conv1d layer, after which fixed sinusoidal positional embeddings encode absolute patch position.
The patch width is a low-impact choice fixed to its default; the supporting sweep is in \cref{sec:supp-arch}.

\paragraph{Transformer blocks.}
The trunk stacks $L_d = 12$ transformer blocks. Each block applies QK-normalised multi-head
self-attention with Rotary Position Embeddings~(RoPE)~\cite{su2024roformer} on
queries/keys; modulates the tokens through AdaLN-Zero~\cite{dit} (per-block scale/shift/gate
derived from~$\mathbf{c}$, gates zero-initialised so each block starts as identity); injects
stimulus structure via a cross-attention layer whose queries are the patch tokens and
keys/values the projected DINOv2 tokens; and ends with a SwiGLU~\cite{shazeer2020glu}
feed-forward network, using RMSNorm~\cite{zhang2019rms} throughout. These components (RoPE,
QK-norm, SwiGLU, RMSNorm, AdaLN-Zero) are standard diffusion-transformer ingredients, not
contributions of this work; our focus is their use within native-space flow matching for fMRI.

\paragraph{Unpatchification.}
A final AdaLN-modulated linear layer maps each patch token back to its corresponding voxel
values, and the patches are concatenated to recover the full-length velocity
prediction~$\hat{\mathbf{u}} \in \mathbb{R}^{L}$.

\subsection{Multi-Subject Shared Trunk with Per-Subject Adapters}
\label{sec:method-ms}

To leverage multiple subjects in one model, StimFlow uses a \emph{shared--private} design: all
transformer blocks, timestep/CLIP embedders, and context projectors are shared, forming a
common trunk that learns subject-invariant stimulus-to-velocity dynamics. The only per-subject
parts are two lightweight adapters at the trunk boundaries: a patch embedder (Conv1d) at the
input and a readout head (AdaLN\,+\,linear) at the output. Since all subjects are zero-padded to
a uniform $L = 2^{14} = 16384$, the patch grid is identical regardless of voxel count;
mini-batches are subject-homogeneous (each forward pass uses a single adapter pair) and a scalar
subject index selects the adapters. This keeps the per-subject parameter count tiny while the
shared trunk benefits from all subjects' data, and, unlike methods that map different subjects'
fMRI into a common latent bottleneck, keeps synthesis in each subject's native voxel space.

\subsection{ROI-Ordered Native Voxel Representation}
\label{sec:method-roi}

Before patchification, voxels are sorted once by functional ROI rather than arbitrary index
(e.g.\ V1--V3, hV4, EBA, FFA, PPA), a static reordering that uses no ROI labels at inference.
Because generation stays in native voxel space, per-ROI fidelity can be read directly off the
synthesized signal by indexing the relevant voxels, without any region-specific decoder or
re-mapping step (\cref{sec:exp-abl,sec:exp-roi-ms}).

\subsection{Training and Inference}
\label{sec:method-train}

\paragraph{Training.}
We optimise the flow-matching loss~(\cref{eq:fm}) with AdamW~\cite{loshchilov2019decoupled}
($\beta_1 = 0.9$, $\beta_2 = 0.95$, weight decay $0.01$) using a cosine learning-rate
schedule~\cite{loshchilov2017sgdr} from $10^{-4}$ to $10^{-5}$ over 50~epochs, with 500 warm-up steps (see \cref{fig:fm}a).
Gradients are clipped to unit norm, and all training is conducted in bf16 mixed
precision.
Effective batch size is $32 \times 8 = 256$ (via gradient accumulation).

\paragraph{Inference.}
At test time, we solve the learned ODE from $t = 0$ (noise) to $t = 1$ (data) with the Euler
method using $N_{\text{step}} = 50$ steps (see \cref{fig:fm}b); each noise sample
$\mathbf{x}_0 \sim \mathcal{N}(\mathbf{0}, \mathbf{I})$ yields a different plausible response.
We report a \emph{single-sample} setting ($T{=}1$) as our primary operating point and an
\emph{optional} multi-sample average ($T{=}N$) approximating the conditional mean. Unlike
diffusion-prior encoders that rely on multi-trial averaging and post-selection, StimFlow
already attains competitive results at $T{=}1$ (\cref{tab:main}); averaging is a deliberate
variance-reduction option, applied identically to every method for fairness.

\enlargethispage{2\baselineskip}\section{Experiments}
\label{sec:exp}

\subsection{Experimental Setup}
\label{sec:exp-setup}

\paragraph{Dataset.}
We use the Natural Scenes Dataset~(NSD)~\cite{nsd}, focusing on the four subjects
(1, 2, 5, 7) who completed all scanning sessions.
Following the standard protocol, each subject has roughly 9{,}000 unique training
images and a shared set of 1{,}000 test images presented up to three times.
We operate on z-scored single-trial betas restricted to the \texttt{nsdgeneral} ROI.

\paragraph{Baselines.}
We compare against two deterministic regression encoders (per-voxel ridge regression and a
transformer encoder trained in our pipeline) and two recent generative fMRI synthesizers,
MindSimulator~\cite{mindsimulator} and SynBrain~\cite{synbrain}. As MindSimulator has no public
implementation, we report its published numbers; SynBrain and both regression baselines are run
in our pipeline. Every method we run uses the same z-scored single-trial betas, the
normalization standard in recent NSD work~\cite{mindsimulator,mindeye2}, and the identical
per-voxel, across-image Pearson metric on the same test split. Re-running SynBrain under this
standardized z-scoring yields lower voxel-level Pearson than its original paper, whose
preprocessing retains the per-voxel mean.

\paragraph{Metrics.}
We evaluate at two levels.
\emph{Voxel-level} metrics (MSE and Pearson correlation) are
computed per voxel across the test images (the standard encoding metric) and averaged over
trials; the Pearson correlation is used consistently in all tables and per-ROI analyses.
\emph{Semantic-level} metrics are obtained by decoding the synthesized fMRI with a
frozen MindEye2~\cite{mindeye2} model and comparing the decoded image to the stimulus
(Inception, CLIP, EfficientNet, SwAV).

\paragraph{Implementation.}
StimFlow and its ablation variants (backbone replacements, single-stream variants) are trained
with the flow-matching objective and schedule of \cref{sec:method-train}. SynBrain is reproduced
from the authors' released code with its original VAE\,+\,S2N objective and hyperparameters;
MindSimulator numbers are from the published paper.

\begin{table}[tb]
  \caption{Visual-to-fMRI synthesis on NSD. StimFlow is the multi-subject model
    (shared trunk with per-subject adapters, trained jointly on subjects 1, 2, 5, 7);
    results are averaged over subjects (per-subject Pearson std $\approx 0.05$;
    per-subject values in \cref{sec:supp-stats}).
    Voxel-level metrics are computed per voxel across images and averaged over trials;
    semantic-level metrics are obtained by decoding the synthesized fMRI with a frozen MindEye2
    model.
    Methods are grouped into deterministic regression baselines, single-sample generative
    ($T{=}1$), and multi-sample generative ($T{=}5$).
    \emph{DiT-1D (MSE)} is an ablative control sharing StimFlow's backbone and DINOv2$+$CLIP
    conditioning but trained with a deterministic MSE loss instead of flow matching, isolating
    the contribution of the generative objective.}
  \label{tab:main}
  \centering
  \setlength{\tabcolsep}{5pt}
  \resizebox{\linewidth}{!}{
  \begin{tabular}{@{}lcccccc@{}}
    \toprule
    & \multicolumn{2}{c}{Voxel-Level} & \multicolumn{4}{c}{Semantic-Level (via decoding)}\\
    \cmidrule(lr){2-3}\cmidrule(lr){4-7}
    Method & MSE$\downarrow$ & Pearson$\uparrow$ & Incep$\uparrow$ & CLIP$\uparrow$
      & Eff$\downarrow$ & SwAV$\downarrow$\\
    \midrule
    \multicolumn{7}{@{}l}{\textit{Regression baselines}}\\
    Linear Regression    & 0.394 & 0.334 & 90.1\% & 87.2\% & 0.728 & 0.432 \\
    Transformer Encoder  & 0.387 & 0.337 & 89.8\% & 85.5\% & 0.759 & 0.440 \\
    DiT-1D (MSE)         & 0.352 & 0.355 & 91.0\% & 89.0\% & 0.720 & 0.412 \\
    \midrule
    \multicolumn{7}{@{}l}{\textit{Generative, single sample ($T{=}1$)}}\\
    MindSimulator        & 0.403 & 0.346 & 92.1\% & 90.4\% & 0.701 & 0.396 \\
    SynBrain             & 0.398 & 0.340 & 90.8\% & 88.5\% & 0.715 & 0.418 \\
    \textbf{StimFlow}    & \textbf{0.342} & \textbf{0.371} & \textbf{94.0\%} & \textbf{91.8\%} & \textbf{0.668} & \textbf{0.375} \\
    \midrule
    \multicolumn{7}{@{}l}{\textit{Generative, multi-sample ($T{=}5$)}}\\
    MindSimulator        & 0.385 & 0.357 & 93.1\% & 91.2\% & 0.689 & 0.391 \\
    SynBrain             & 0.388 & 0.352 & 91.5\% & 89.4\% & 0.705 & 0.410 \\
    \textbf{StimFlow}    & \textbf{0.335} & \textbf{0.387} & \textbf{95.0\%} & \textbf{92.8\%} & \textbf{0.650} & \textbf{0.362} \\
    \midrule
    GT fMRI (single-trial) & --   & --    & 96.9\% & 93.9\% & 0.604 & 0.342 \\
    \bottomrule
  \end{tabular}}
\end{table}

\begin{table}[tb]
  \caption{Feature backbone ablation, averaged over subjects 1, 2, 5, 7.
    The default StimFlow conditions on DINOv2 (structural, via cross-attention) and
    CLIP (semantic, via AdaLN).
    Each alternative backbone replaces \emph{both} feature streams.
    ``DINOv2 only'' and ``CLIP only'' retain a single original stream.}
  \label{tab:abl-backbone}
  \centering
  \setlength{\tabcolsep}{6pt}
  \begin{tabular}{@{}lcccc@{}}
    \toprule
    & \multicolumn{2}{c}{$T{=}1$} & \multicolumn{2}{c}{$T{=}5$}\\
    \cmidrule(lr){2-3}\cmidrule(lr){4-5}
    Feature backbone & MSE$\downarrow$ & Pearson$\uparrow$ & MSE$\downarrow$ & Pearson$\uparrow$\\
    \midrule
    ResNet-50        & 0.354 & 0.348 & 0.346 & 0.364 \\
    ConvNeXt         & 0.355 & 0.345 & 0.347 & 0.361 \\
    ViT-B/16         & 0.347 & 0.363 & 0.339 & 0.379 \\
    \midrule
    DINOv2 only      & 0.354 & 0.351 & 0.346 & 0.368 \\
    CLIP only        & 0.361 & 0.338 & 0.354 & 0.351 \\
    \midrule
    \textbf{DINOv2 + CLIP} (default) & \textbf{0.342} & \textbf{0.371} & \textbf{0.335} & \textbf{0.387} \\
    \bottomrule
  \end{tabular}
\end{table}

\subsection{Main Results: Visual-to-fMRI Synthesis}
\label{sec:exp-main}

\Cref{tab:main} shows StimFlow improving over every baseline on all six metrics at both
$T{=}1$ and $T{=}5$.
To isolate the contribution of the flow-matching objective from the rest of our design,
we add a \emph{DiT-1D (MSE)} control: the identical backbone and DINOv2$+$CLIP conditioning,
trained with a deterministic per-voxel MSE regression loss instead of conditional flow
matching. This control already reaches per-voxel Pearson $0.355$ (MSE $0.352$), above both
generative baselines, which indicates that most of our gain comes from the native-space
formulation and the dual conditioning rather than from the generative objective per se.
Replacing the MSE loss with conditional flow matching (default StimFlow, $T{=}1$) adds a
further $+0.016$ in voxel Pearson ($0.355{\rightarrow}0.371$) and a comparable improvement
on the semantic metrics; the more decisive benefit of flow matching is its
train--inference alignment, which lets a single sample ($T{=}1$) match what the generative
baselines require five draws to achieve (see below).
At the voxel level, a single sample reaches Pearson $0.371$, a consistent advantage
over MindSimulator, with MSE $0.342$.
Per-subject bootstrap confidence intervals ($B{=}10{,}000$) are tight (width ${\approx}\,0.007$;
e.g.\ subject 1: $0.375\,[0.372,\,0.378]$), so each within-subject estimate is precise;
the cross-subject standard deviation ($0.051$) reflects individual variability (subject~7
has the lowest noise ceiling), not estimation noise. Because $n=4$ limits formal
statistical testing, we report this as a consistent directional advantage across subjects.
More saliently, StimFlow at $T{=}1$ already matches or exceeds every baseline at $T{=}5$ on both
voxel metrics, indicating the formulation reduces the need for multi-sample averaging.
At the semantic level, decoding StimFlow responses with the frozen MindEye2 approaches the GT
upper bound (Inception $94.0\%$ vs.\ $96.9\%$; CLIP $91.8\%$ vs.\ $93.9\%$ at $T{=}1$), and
averaging lifts Inception to $95.0\%$, within two points of GT.

\subsection{Backbone Ablation and Feature Complementarity}
\label{sec:exp-abl}

\begin{figure}[!ht]
  \centering
  \includegraphics[width=0.8\linewidth]{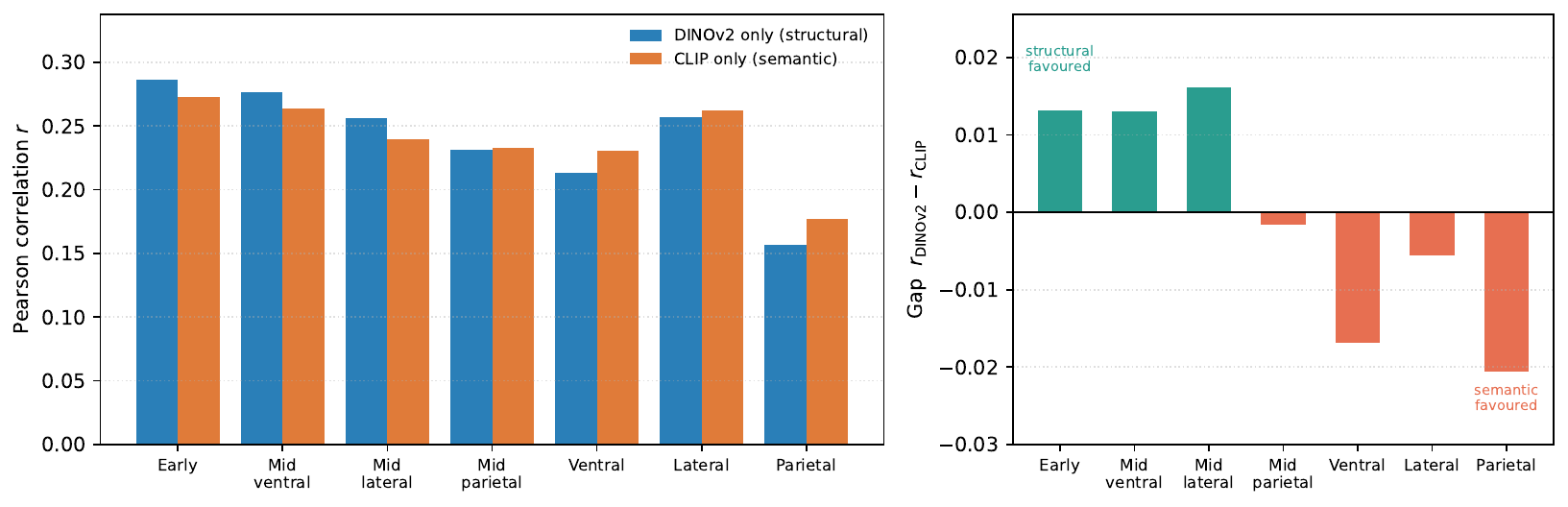}
  \caption{\textbf{Per-stream feature complementarity}, averaged over subjects 1, 2, 5, 7.
    (a)~Per-voxel Pearson correlation of the DINOv2-only (structural) and CLIP-only (semantic)
    conditioning variants, by NSD functional stream (early\,$\rightarrow$\,higher).
    (b)~Their gap ($r_{\mathrm{DINOv2}}-r_{\mathrm{CLIP}}$): on average the structural
    stream leads in early-to-mid visual cortex and the advantage reverses in the higher
    streams (Spearman $\rho = -0.86$, $p < 0.05$ across the seven streams),
    suggesting a structure-to-semantics gradient.}
  \label{fig:roi-gap}
\end{figure}
Conditioning on a structural (DINOv2) and a semantic (CLIP) stream is a deliberate,
empirically grounded choice. We justify it with two experiments: a feature-backbone ablation
asking \emph{which} features to condition on (\cref{tab:abl-backbone}), and a per-ROI analysis
asking \emph{why} the two streams work better together (\cref{fig:roi-gap}). The ablation
replaces both conditioning streams with standard supervised backbones
(ResNet-50~\cite{he2016deep}, ConvNeXt~\cite{Liu_2022_CVPR}, ViT-B/16~\cite{dosovitskiy2021image})
and also retains each stream in isolation, at both $T{=}1$ and $T{=}5$.

\paragraph{The pairing is deliberate (\cref{tab:abl-backbone}).}
Three observations show the choice is not arbitrary.
First, the conditioning backbone matters substantially even with the synthesis network
fixed: $T{=}1$ per-voxel Pearson ranges from $0.345$ (ConvNeXt) to $0.371$ (default),
a spread of $0.026$ comparable to the gaps among methods in \cref{tab:main}.
Notably, even a generic ViT-B/16 backbone ($0.363$) already surpasses both generative
baselines (MindSimulator $0.346$, SynBrain $0.340$), reinforcing the conclusion from
\cref{sec:exp-main} that the gain stems primarily from native-space synthesis and dual
conditioning rather than the conditioning backbone alone; consistent with the
DiT-1D (MSE) control, the flow-matching objective adds a further, more modest voxel-level
improvement on top.
Second, each original stream in isolation (DINOv2-only $0.351$, CLIP-only $0.338$)
falls below the ViT-B/16 single-backbone result; the dual default recovers the lead
($0.371$), but the aggregate margin over ViT-B/16 is modest ($+0.008$) and should not
be over-interpreted given four subjects.
The stronger evidence for dual-stream conditioning is \emph{mechanistic} rather than
aggregate: as the per-ROI analysis below shows, each stream dominates in a different
part of the cortical hierarchy, and prior generative encoders that condition
\emph{exclusively} on CLIP sacrifice coverage of early visual cortex.
That ViT-B/16 alone outperforms both specialist single streams suggests the key
requirement is conditioning that spans the full hierarchy; DINOv2+CLIP is one effective
and interpretable instantiation of this principle.

\paragraph{The per-ROI breakdown suggests complementarity
(\cref{fig:roi-gap}).}
To understand \emph{why} the pairing helps, \cref{fig:roi-gap} compares the two single-stream
variants per NSD functional stream ($T{=}1$ per-voxel Pearson, averaged over subjects).
The DINOv2-only advantage is positive in early-to-mid streams ($+0.013$ to $+0.016$
on average) and turns negative in the higher ventral, lateral, and parietal streams
(down to $-0.021$), suggesting a structure-to-semantics gradient along the hierarchy.
Across the seven functional streams, the Spearman rank correlation between hierarchical
position and the DINOv2\,--\,CLIP gap is $\rho = -0.86$ ($p < 0.05$), and all four
subjects individually show a negative trend ($\rho$ ranging from $-0.21$ to $-0.93$).
The per-ROI differences are modest and limited to four subjects, so the crossover should be read
as a \emph{descriptive trend consistent with known functional specialisation} rather than a
definitive per-region claim; nonetheless, structural features being more informative in early
cortex and semantic features in higher areas offers a plausible account of why dual conditioning
helps.

\subsection{Per-ROI Synthesis Fidelity of the Multi-Subject Model}
\label{sec:exp-roi-ms}
\begin{figure}[!ht]
  \centering
  \includegraphics[width=0.7\linewidth]{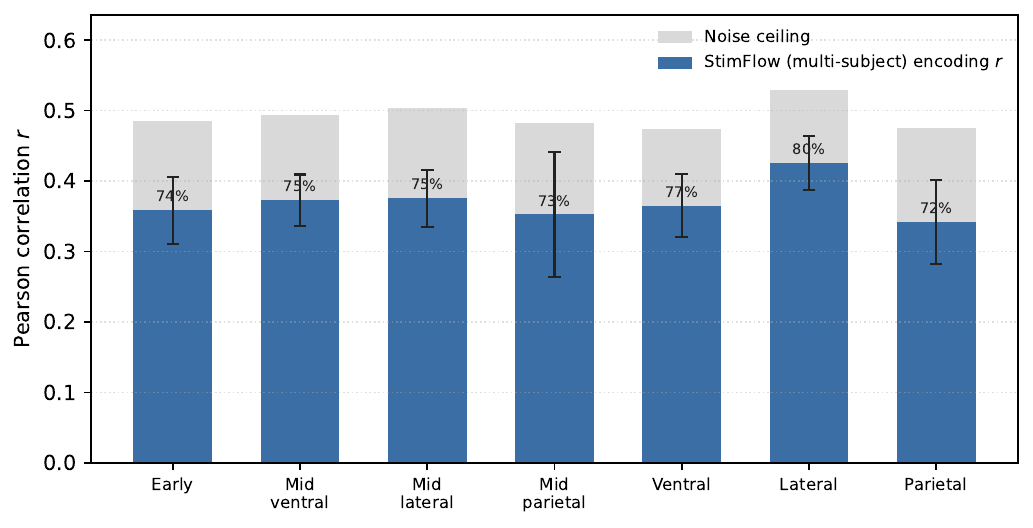}
  \caption{\textbf{Per-stream fidelity of the multi-subject model.} Per-voxel Pearson
    correlation of multi-subject StimFlow (shared trunk + per-subject adapters),
    mean\,$\pm$\,std over subjects 1/2/5/7, overlaid on the per-stream
    noise ceiling (grey); percentages give the captured fraction of the ceiling. Fidelity
    remains a substantial fraction of the ceiling across the visual hierarchy, with
    no region showing a pronounced drop.}
  \label{fig:roi-ms}
\end{figure}

We next verify that the deployed multi-subject model (shared trunk with per-subject adapters,
jointly trained on subjects 1, 2, 5, 7) preserves region-specific structure. \Cref{fig:roi-ms}
reports its per-stream fidelity as the per-voxel Pearson correlation relative to the per-stream
noise ceiling. Accuracy remains stable across the hierarchy and recovers a substantial fraction
of the ceiling in every stream, with no pronounced weak region, read directly off the generated
signal, since synthesis stays in native voxel space.

\subsection{Cross-Subject Generalization and Few-Shot Adaptation}
\label{sec:exp-ms}
Finally, we evaluate cross-subject generalization under a leave-one-subject-out protocol. The shared trunk is trained on three subjects and adapted to the held-out one by fitting only its per-subject input/output adapters and a low-rank (LoRA) residual~\cite{hu2022lora} on the last block (${\sim}0.62$\,M trainable parameters, under $1\%$ of the trunk). Scaling the held-out subject's data from 1 to 4 hours (one hour ${\approx}\,750$ trials), \cref{tab:fewshot} reports the per-voxel Pearson correlation and the fraction of the full-data reference recovered, for $T{=}1$ and $T{=}5$; the reference is the multi-subject model trained on all four subjects with full ${\sim}40$\,h each (\cref{tab:main,tab:abl-backbone}). At $T{=}5$, adaptation recovers ${\sim}60\%$ from one hour and ${\sim}69\%$ at 4\,h; even $T{=}1$ recovers ${\sim}54\%$ at 1\,h, showing the shared trunk transfers to unseen subjects with limited data.

\begin{table}[tb]
  \caption{Few-shot novel-subject adaptation with increasing data budget under a
    \emph{leave-one-subject-out} protocol. The shared trunk is trained on three subjects and adapted
    to the held-out subject (only its per-subject adapters + LoRA) using $1$--$4$\,h of that
    subject's stimulus data. We report per-voxel Pearson correlation (per voxel, across images)
    for both single-sample ($T{=}1$) and
    multi-sample ($T{=}5$) synthesis. The \emph{Full} column is the
    multi-subject StimFlow model trained on all four subjects with full data
    (${\sim}40$\,h each); parenthesised values give the fraction of
    that full-data reference recovered.}
  \label{tab:fewshot}
  \centering
  \setlength{\tabcolsep}{5pt}
  \renewcommand{\arraystretch}{1.15}
  \resizebox{\linewidth}{!}{
  \begin{tabular}{@{}llcccccc@{}}
    \toprule
    Held-out & Train & & Full & \multicolumn{4}{c}{Few-shot budget}\\
    \cmidrule(lr){5-8}
    subject & set & Trials & (ref.) & 1\,h & 2\,h & 3\,h & 4\,h\\
    \midrule
    \multirow{2}{*}{Sub1} & \multirow{2}{*}{2,5,7}
      & $T{=}1$ & 0.375\,(100\%) & 0.182\,(48\%) & 0.189\,(50\%) & 0.206\,(55\%) & 0.210\,(56\%)\\
    & & $T{=}5$ & 0.391\,(100\%) & 0.215\,(55\%) & 0.225\,(58\%) & 0.249\,(64\%) & 0.251\,(64\%)\\
    \midrule
    \multirow{2}{*}{Sub2} & \multirow{2}{*}{1,5,7}
      & $T{=}1$ & 0.384\,(100\%) & 0.193\,(50\%) & 0.208\,(54\%) & 0.227\,(59\%) & 0.228\,(59\%)\\
    & & $T{=}5$ & 0.398\,(100\%) & 0.234\,(59\%) & 0.255\,(64\%) & 0.268\,(67\%) & 0.271\,(68\%)\\
    \midrule
    \multirow{2}{*}{Sub5} & \multirow{2}{*}{1,2,7}
      & $T{=}1$ & 0.423\,(100\%) & 0.260\,(61\%) & 0.271\,(64\%) & 0.281\,(66\%) & 0.290\,(68\%)\\
    & & $T{=}5$ & 0.437\,(100\%) & 0.294\,(67\%) & 0.314\,(72\%) & 0.326\,(74\%) & 0.331\,(76\%)\\
    \midrule
    \multirow{2}{*}{Sub7} & \multirow{2}{*}{1,2,5}
      & $T{=}1$ & 0.301\,(100\%) & 0.166\,(55\%) & 0.170\,(56\%) & 0.176\,(58\%) & 0.175\,(58\%)\\
    & & $T{=}5$ & 0.320\,(100\%) & 0.193\,(60\%) & 0.206\,(64\%) & 0.212\,(66\%) & 0.213\,(67\%)\\
    \midrule
    \multirow{2}{*}{\textbf{Average}} & \multirow{2}{*}{--}
      & $T{=}1$ & \textbf{0.371\,(100\%)} & \textbf{0.200\,(54\%)} & \textbf{0.210\,(57\%)}
        & \textbf{0.222\,(60\%)} & \textbf{0.226\,(61\%)}\\
    & & $T{=}5$ & \textbf{0.387\,(100\%)} & \textbf{0.234\,(60\%)} & \textbf{0.250\,(65\%)}
      & \textbf{0.264\,(68\%)} & \textbf{0.267\,(69\%)}\\
    \bottomrule
  \end{tabular}}
\end{table}

The recovered fraction (\cref{tab:fewshot}) rises steadily with budget for every held-out
subject, steepest in the first hour, still gaining through $4$\,h rather than saturating, and
multi-sample averaging adds a further $6$--$8$ points. That one hour recovers over half the
full-data reference by adapting only two per-subject linear layers and a low-rank residual
indicates the shared trunk captures much of the transferable visual-to-neural structure.

\subsection{Qualitative Analysis}
\label{sec:exp-qual}

\begin{figure}[!ht]
  \centering
  \includegraphics[width=0.9\linewidth]{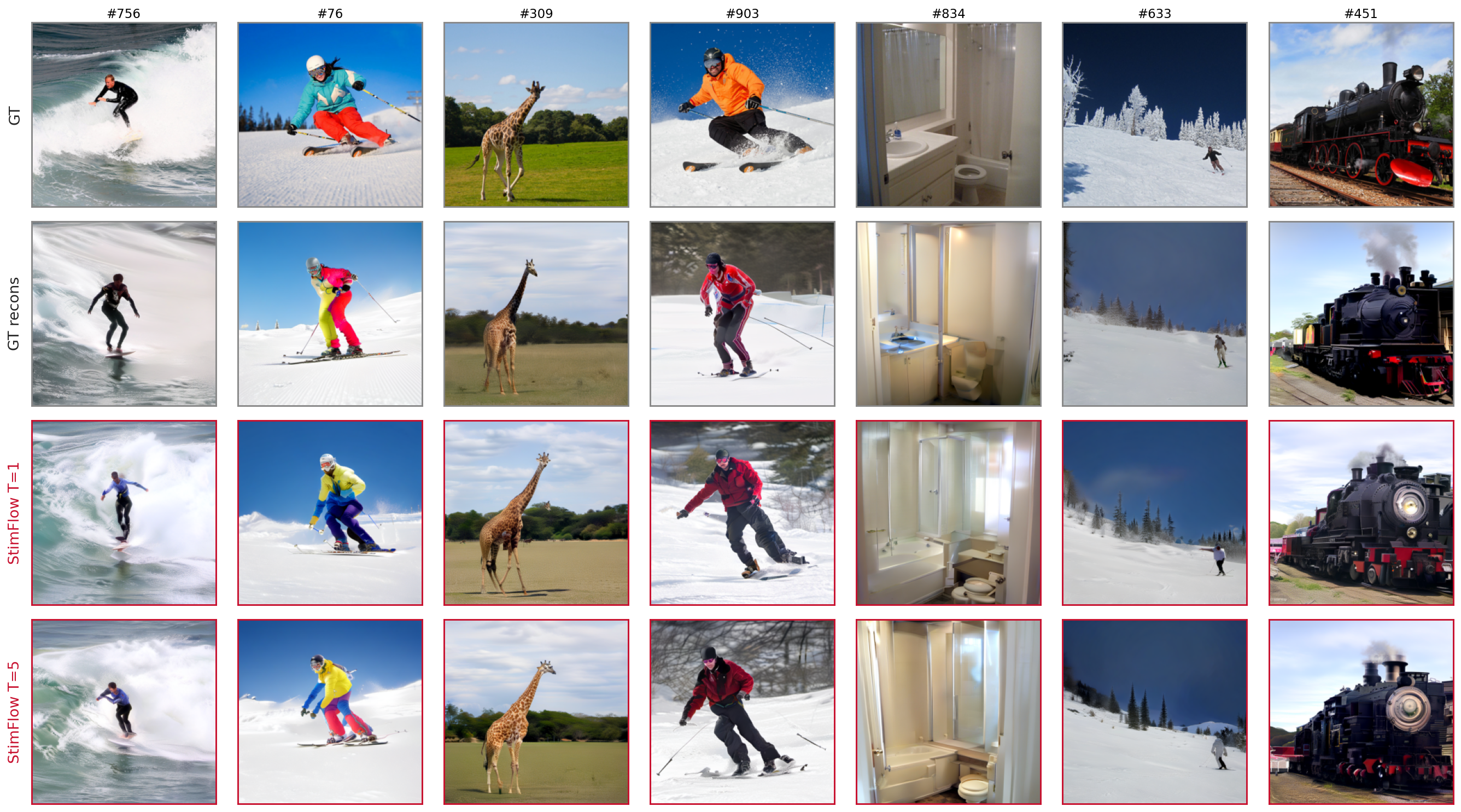}
  \caption{\textbf{Qualitative reconstruction comparison (subject 1).} Each column is one
    test image. Rows, top to bottom: the seen stimulus (GT), the MindEye2 reconstruction
    from ground-truth fMRI, and the reconstructions from StimFlow-synthesized fMRI at
    $T{=}1$ and $T{=}5$. Reconstructions from synthesized fMRI recover the same semantic
    content as those from real fMRI.}
  \label{fig:recon}
\end{figure}

\paragraph{Reconstruction comparison.}
\Cref{fig:recon} shows the semantic-level evaluation qualitatively: reconstructions decoded
(via frozen MindEye2) from StimFlow-synthesized versus real fMRI. Across diverse content
(a surfer, skiers, a giraffe, a steam locomotive), the synthesized reconstructions recover
object identity, pose, layout, and dominant colour. As expected, real-fMRI reconstructions
remain strongest and bound what any synthesized signal can achieve under the same decoder.
Among synthesized variants, $T{=}5$ is clearly better (cleaner, sharper), while $T{=}1$ recovers
category and coarse layout but is noisier, already capturing the dominant semantics that
averaging sharpens rather than introduces. The hardest case is the indoor bathroom scene, whose
fine spatial layout is only partially recovered, in line with the general difficulty of
fine-grained spatial structure relative to dominant-category content.

\begin{figure}[!ht]
  \centering
  \includegraphics[width=0.9\linewidth]{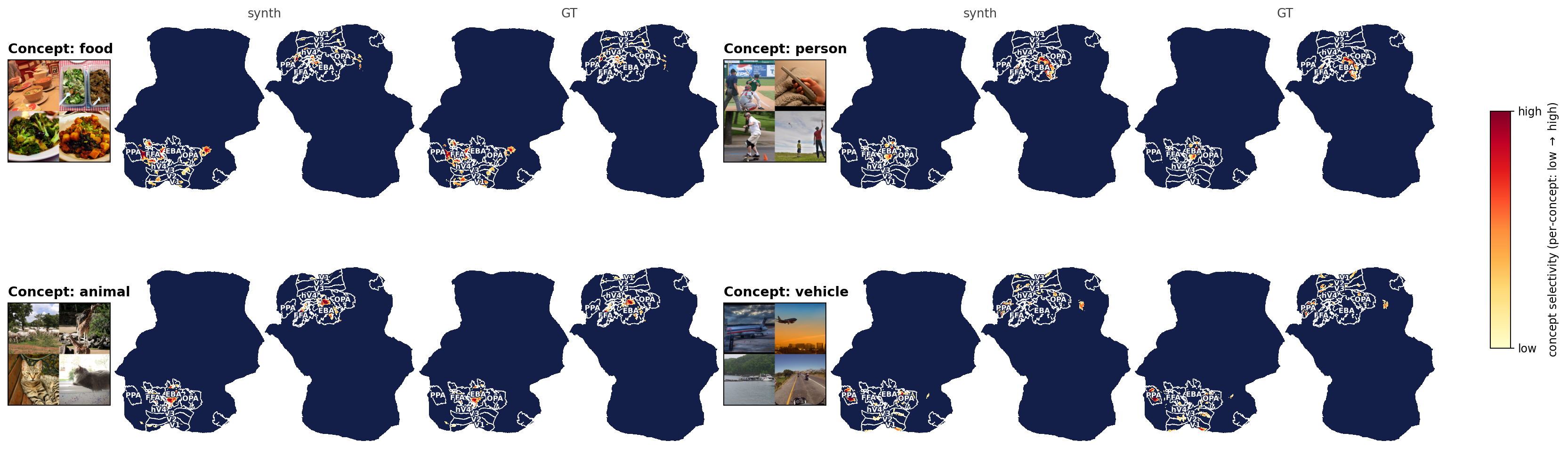}
  \caption{\textbf{Concept-selective region localization (subject~1).} For each concept
    (rows), a montage of example stimuli is shown beside the selectivity map (mean response
    over the concept's images minus the average response across all test images) rendered on
    the cortical flatmap,
    from StimFlow's synthesized responses (\emph{synth}) and the ground-truth responses
    (\emph{GT}). White contours mark functional ROIs (V1--hV4, FFA/OFA, PPA/OPA, EBA, VWFA).
    Each concept is scaled to its own range. The synthesized maps localize the same
    concept-selective regions as the ground truth.}
  \label{fig:concept}
\end{figure}

\paragraph{Concept-selective region localization.}
\label{sec:exp-concept}
A key promise of generative encoding is localizing concept-selective cortex from synthesized
responses without new scans~\cite{mindsimulator}. Grouping the shared test images by caption
concept, we average the synthesized responses per group and subtract the all-image mean to
form a per-voxel selectivity map on the subject-1 flatmap (\cref{fig:concept}). The four
concepts produce \emph{distinct}, anatomically sensible signatures: \emph{person} in
body/face-responsive cortex (EBA, FFA), \emph{vehicle} in scene/place regions (PPA, OPA), and
\emph{food}/\emph{animal} in ventral-temporal cortex near FFA/hV4. The synthesized maps place
their peaks in the same ROIs as the ground truth, remaining sparse and focal with only minor
differences in extent.
Quantitatively, the Pearson correlation between synthesized and ground-truth selectivity maps
(per concept across all voxels, averaged over concepts and subjects) is $r = 0.94$ (range
$0.88$--$0.98$) with a top-$10\%$ Dice overlap of $0.81$, confirming the agreement holds
quantitatively, not only by impression.

\section{Conclusion}
\label{sec:conclusion}

Three findings emerge from our experiments.
First, the primary practical benefit of the generative formulation is
\emph{train--inference alignment}: because the flow-matching path begins at the same
Gaussian distribution used at test time, a single sample already matches or exceeds
multi-sample baselines, reducing the need for post-hoc averaging or selection.
The model does \emph{not} yet capture meaningful trial-to-trial variability:
averaging further improves every metric, indicating that the inter-sample variance is
estimation noise rather than neural signal.
Single-sample fidelity remains high precisely because training and inference share the
same starting distribution.
Second, generating in native voxel space, rather than in a compressed latent code,
unlocks analyses that prior generative encoders cannot perform directly: per-ROI fidelity
readout, concept-selective region localization, and cross-subject adaptation via
lightweight adapters, all without an auxiliary decoder.
Third, the structural (DINOv2) and semantic (CLIP) conditioning streams show a
consistent complementary trend along the cortical hierarchy ($\rho = -0.86$,
$p < 0.05$), and their pairing yields gains over either stream alone while
matching or slightly exceeding generic backbone replacements; the aggregate margin
is modest, but the per-ROI mechanistic rationale provides a principled basis for
the dual-stream design.

\paragraph{Limitations and future work.}
StimFlow approximates the conditional mean of $p(\mathbf{v}\mid\mathbf{x})$; it does not
yet model the structured component of trial-to-trial variability (e.g.\ cognitive-state
or attention-dependent fluctuations), which remains an open direction.
Performance is bounded by frozen backbones (${\sim}72$--$80\%$ noise ceiling), and
validation is limited to four NSD subjects; broader evaluation across datasets, tasks,
and populations is needed before generated responses could supplement,
let alone substitute for, empirical data in applied or clinical settings.

\bibliographystyle{splncs04}
\bibliography{main}
\clearpage
\appendix
\renewcommand{\thesection}{S\arabic{section}}
\renewcommand{\thetable}{S\arabic{table}}
\renewcommand{\thefigure}{S\arabic{figure}}
\setcounter{section}{0}\setcounter{table}{0}\setcounter{figure}{0}
\section*{Supplementary Material}
\noindent
This supplementary material reports the design sweeps that justify the
hyperparameters of StimFlow. We separate every choice into \emph{high-impact}
settings, reported in the main text because they shape fidelity, and
\emph{low-impact} settings, fixed to robust defaults and tabulated here so the
main architecture description stays focused. Four axes are covered: the DiT-1D
architecture (\cref{sec:supp-arch}), the training loss and time schedule
(\cref{sec:supp-loss}), the inference solver budget (\cref{sec:supp-nfe}), and
the few-shot LoRA adapter (\cref{sec:supp-lora}).

\section{Protocol}
\label{sec:supp-protocol}
Each sweep varies a \emph{single} hyperparameter while holding all others at
their default values (patch width $P{=}32$, token width $D{=}512$, attention
heads $H{=}8$, depth $L_d{=}12$, sequence length $L{=}16384$). All variants use
the default dual-stream conditioning (DINOv2~$+$~CLIP) and the multi-subject
shared-trunk model trained jointly on subjects 1, 2, 5, and 7 with the optimisation
schedule of the main text (\S3.7). We report voxel-level MSE and Pearson
correlation (per voxel, across images) at the single-sample operating point
($T{=}1$), averaged over the four subjects; the default configuration reproduces
the main-paper number (Pearson $0.371$, MSE $0.342$). Parameter counts are for the
deployed multi-subject model: the shared trunk ($71.9$\,M at the default setting)
plus four per-subject adapter pairs ($\approx0.56$\,M each).

\section{DiT-1D architecture}
\label{sec:supp-arch}

\paragraph{Low-impact: patch width and attention heads (\cref{tab:supp-arch}).}
Both are minor relative to depth and are therefore omitted from the main text. Head
count only repartitions a fixed token width and is essentially constant (within
$0.004$ Pearson). Patch width trades token count (hence self-attention cost) against
voxel granularity and stays flat near the default ($P{=}16$ vs.\ $32$ within $0.003$),
dropping only at the coarsest setting ($P{=}64$, $-0.014$); this confirms that the
per-ROI readout of the main text does \emph{not} rely on any patch-to-region alignment.
Both leave the parameter count unchanged ($\approx74$\,M), so the defaults are
robust rather than tuned.

\begin{table}[h]
  \centering
  \caption{Low-impact architecture sweeps ($T{=}1$, averaged over subjects
    1/2/5/7). Each block varies one hyperparameter with the rest at default;
    the default row (MSE $0.342$, Pearson $0.371$) is the same model, repeated as
    the anchor for each axis. $N$ is the token count, $d$ the per-head dimension.}
  \label{tab:supp-arch}
  \setlength{\tabcolsep}{8pt}
  \begin{tabular}{@{}llcc@{}}
    \toprule
    Hyperparameter & Value & MSE$\downarrow$ & Pearson$\uparrow$\\
    \midrule
    \multirow{3}{*}{Patch width $P$}
      & $16$ ($N{=}1024$)        & 0.341 & 0.368 \\
      & \textbf{$32$ (default)}  & \textbf{0.342} & \textbf{0.371} \\
      & $64$ ($N{=}256$)         & 0.349 & 0.357 \\
    \midrule
    \multirow{3}{*}{Heads $H$}
      & $4$ ($d{=}128$)          & 0.344 & 0.367 \\
      & \textbf{$8$ (default)}   & \textbf{0.342} & \textbf{0.371} \\
      & $16$ ($d{=}32$)          & 0.343 & 0.369 \\
    \bottomrule
  \end{tabular}
\end{table}

\paragraph{High-impact: trunk depth (\cref{tab:supp-depth}).}
Depth is the one architecture knob with a first-order effect: $6\!\rightarrow\!12$
blocks adds $0.019$ Pearson, after which accuracy plateaus ($12\!\rightarrow\!24$
adds $\le 0.001$ at $2\times$ the parameters). We therefore expose $L_d{=}12$ in the
main text (the depth at which the trunk is expressive enough without
over-parameterising).

\begin{table}[h]
  \centering
  \caption{Trunk-depth sweep ($T{=}1$, averaged over subjects 1/2/5/7). Default in bold.}
  \label{tab:supp-depth}
  \setlength{\tabcolsep}{8pt}
  \begin{tabular}{@{}lccc@{}}
    \toprule
    Depth $L_d$ & Params (M) & MSE$\downarrow$ & Pearson$\uparrow$\\
    \midrule
    $6$         & 39.4  & 0.351 & 0.352 \\
    \textbf{$12$ (default)} & 74.1 & \textbf{0.342} & \textbf{0.371} \\
    $18$        & 108.8 & 0.341 & 0.372 \\
    $24$        & 143.4 & 0.342 & 0.371 \\
    \bottomrule
  \end{tabular}
\end{table}

\section{Loss and time-schedule settings}
\label{sec:supp-loss}
For completeness we record the exact settings behind the flow-matching objective
(\cref{eq:fm}); these are standard choices that we do not tune. The training
time $t$ follows a logit-normal schedule,
$\mathrm{logit}(t)\sim\mathcal{N}(\mu,\sigma^2)$ with $\mu{=}-1$ and $\sigma{=}1$,
warped by the resolution-aware shift $t\!\mapsto\! S\,t/\big(1+(S{-}1)\,t\big)$ with
$S$ tied to the sequence length. The per-voxel weight $w_j$ is the voxel's
noise-ceiling reliability, normalised to unit mean over the real (non-padded)
voxels, and the loss is masked to those voxels (both as defined in the main text).

\section{Inference solver budget}
\label{sec:supp-nfe}
\begin{figure}[!ht]
  \vspace{-3mm}
  \centering
  \begin{tikzpicture}
    \begin{axis}[
      width=0.72\linewidth, height=4.0cm,
      xlabel={Solver steps},
      ylabel={Pearson $r$ ($T{=}1$)},
      xmin=0, xmax=110, ymin=0.335, ymax=0.380,
      xtick={10,25,50,100},
      ytick={0.34,0.35,0.36,0.37},
      grid=both, grid style={gray!18},
      legend pos=south east, legend cell align=left,
    ]
      \addplot[mark=*, thick, blue] coordinates
        {(10,0.340) (25,0.365) (50,0.371) (100,0.372)};
      \addlegendentry{Euler (1 NFE/step)}
      \addplot[only marks, mark=star, mark size=3.5pt, red] coordinates {(25,0.372)};
      \addlegendentry{Heun (2 NFE/step)}
      \draw[dashed, gray] (axis cs:50,0.335) -- (axis cs:50,0.371);
      \node[anchor=south, font=\scriptsize, gray] at (axis cs:50,0.375)
        {default};
      \node[anchor=west, font=\scriptsize, red!70!black] at (axis cs:27,0.372)
        {$25$ steps ($50$ NFE)};
    \end{axis}
  \end{tikzpicture}
  \caption{Inference solver vs.\ step count ($T{=}1$, averaged over subjects
    1/2/5/7). Euler fidelity (blue) plateaus by $50$ steps (default, dashed line).
    Heun (red star) reaches the same plateau in $25$ steps, but requires the
    same $50$ NFEs as Euler-$50$ ($2$ NFEs/step), with no gain at equal compute.}
  \label{fig:supp-nfe}
  \vspace{-3mm}
\end{figure}
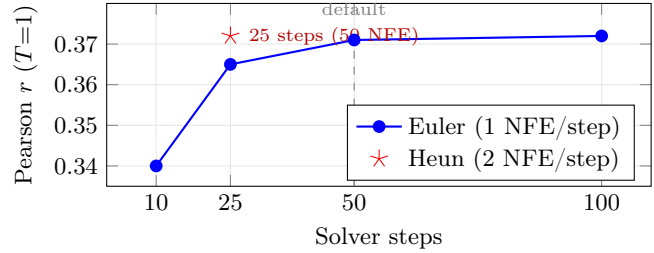

At test time we integrate the learned ODE with a fixed-step solver
(\cref{fig:supp-nfe}). Euler fidelity saturates by $50$ steps; halving to $25$
costs only $0.006$ Pearson while doubling to $100$ adds nothing. A second-order
Heun solver reaches the same plateau in $25$ steps, but since it spends two network
evaluations per step this is the \emph{same} $50$ function evaluations as
Euler-$50$, i.e.\ no gain at equal compute. We therefore report the simpler
Euler, $50$-step setting in the main text and treat the solver as a non-sensitive
inference knob. This sweep is inference-only: all points reuse the single default
checkpoint.

\section{Few-shot LoRA adaptation}
\label{sec:supp-lora}
The few-shot protocol (\cref{sec:exp-ms}) adapts the frozen shared trunk to a
held-out subject with its per-subject adapters plus a LoRA residual on the last
trunk block. \Cref{tab:supp-lora} sweeps the LoRA rank and the number of adapted
trailing blocks at the $1$\,h budget ($T{=}5$, averaged over the four held-out
subjects). Adapters alone (no LoRA) leave the frozen trunk unable to specialise; a
rank-$8$ residual on a single block recovers most of the gain, after which larger
rank or more blocks add little for roughly double the trainable parameters. We thus
fix rank $8$ on the last block, adapting only its attention and feed-forward Linear
layers while norms and AdaLN conditioning stay frozen.

\begin{table}[!ht]
  \vspace{-3mm}
  \centering
  \caption{LoRA rank and adapted-block sweep for few-shot adaptation
    ($1$\,h budget, $T{=}5$, averaged over the four held-out subjects).
    ``Trainable'' counts the per-subject adapters ($0.56$\,M) plus the LoRA
    residual. Default in bold.}
  \label{tab:supp-lora}
  \setlength{\tabcolsep}{7pt}
  \begin{tabular}{@{}llcc@{}}
    \toprule
    Adapted blocks & Rank & Trainable (M) & Pearson$\uparrow$\\
    \midrule
    none (adapters only) & --   & 0.559 & 0.204 \\
    last $1$             & $4$  & 0.592 & 0.226 \\
    \textbf{last $1$}    & \textbf{$8$}  & \textbf{0.625} & \textbf{0.234} \\
    last $1$             & $16$ & 0.690 & 0.235 \\
    last $2$             & $8$  & 0.690 & 0.236 \\
    \bottomrule
  \end{tabular}
  \vspace{-3mm}
\end{table}

\section{Per-subject voxel-level statistics}
\label{sec:supp-stats}

\Cref{tab:supp-persubj} reports the per-subject per-voxel Pearson correlation
of the default StimFlow multi-subject model (shared trunk + per-subject adapters)
together with bootstrap 95\% confidence intervals ($B{=}10{,}000$).
Within each subject the estimate is highly precise (CI width ${\approx}\,0.007$),
confirming that the voxel-level metric is not driven by estimation noise.
The cross-subject standard deviation ($0.051$) reflects genuine individual
differences in data quality and noise ceiling: subject~5 has the highest
noise ceiling and the best performance, while subject~7 has the lowest.

Multi-sample averaging ($T{=}5$) significantly improves over single-sample
($T{=}1$) within every subject (paired bootstrap $p < 10^{-4}$,
$\Delta \approx +0.016$ per subject, $B{=}10{,}000$).

\begin{table}[h]
  \centering
  \caption{Per-subject voxel-level Pearson correlation (mean over voxels,
    computed across images) with bootstrap 95\% CI ($B{=}10{,}000$).
    Default multi-subject StimFlow model.}
  \label{tab:supp-persubj}
  \setlength{\tabcolsep}{6pt}
  \begin{tabular}{@{}lccc@{}}
    \toprule
    Subject & $N_{\text{voxels}}$ & $T{=}1$\;Pearson\;[95\% CI] & $T{=}5$\;Pearson\;[95\% CI]\\
    \midrule
    1 & 15\,724 & $0.375\;[0.372,\;0.378]$ & $0.391\;[0.388,\;0.394]$\\
    2 & 14\,278 & $0.384\;[0.380,\;0.388]$ & $0.398\;[0.394,\;0.402]$\\
    5 & 13\,039 & $0.423\;[0.419,\;0.427]$ & $0.437\;[0.434,\;0.441]$\\
    7 & 12\,682 & $0.301\;[0.297,\;0.304]$ & $0.320\;[0.317,\;0.323]$\\
    \midrule
    Average & -- & $0.371\;\pm 0.051$ & $0.387\;\pm 0.049$\\
    \bottomrule
  \end{tabular}
\end{table}

\section{The SynBrain Baseline: Why Its Reported Per-Image Pearson Is Inflated}
\label{sec:supp-norm}

This section explains why SynBrain's originally reported Pearson ($0.687$) is far above the
comparable value ($0.340$). The effect is specific to the \emph{per-image, across-voxel}
Pearson (the metric SynBrain reports) combined with its normalization, and is therefore
analyzed here in that metric; it is distinct from the per-voxel, across-image encoding Pearson
used for all results in the main text.

\paragraph{Two normalization regimes.}
The main text z-scores the single-trial betas: each voxel is centred and scaled to zero mean and
unit variance using training-set statistics, following the standard NSD encoding/decoding
protocol~\cite{mindsimulator,mindeye2,nsd}. The publicly released SynBrain pipeline instead
divides every beta by a fixed constant ($2000$) applied uniformly to all voxels, \emph{without}
removing the per-voxel mean. We denote the two regimes \emph{mean-retained} and \emph{z-scored}.

\paragraph{SynBrain's metric, and why the normalization matters.}
SynBrain scores fidelity with the per-image, across-voxel Pearson correlation: for test image $i$
it correlates the synthesized voxel pattern $\hat{\mathbf v}_i\in\mathbb{R}^{V}$ against the
measured pattern $\mathbf v_i$ across the $V$ voxels, then averages over images. Crucially, this
correlation is invariant to any positive rescaling,
$\mathrm{PCC}(a\hat{\mathbf v}_i,\mathbf v_i)=\mathrm{PCC}(\hat{\mathbf v}_i,\mathbf v_i)$ for
$a>0$, so the constant division by $2000$ \emph{by itself cannot change this metric}. The
regimes differ in two ways --- z-scoring removes the per-voxel mean and additionally rescales
each voxel by its own standard deviation --- but, as the mean-only control below isolates, it is
the retained \emph{per-voxel mean}, not the rescaling, that drives the inflation: under
across-voxel scoring with the mean kept, a model only has to reproduce the static per-voxel mean
pattern to attain a high score, so the reported Pearson is largely an artifact of the
normalization rather than evidence of stimulus-specific fidelity.

\paragraph{Why mean retention inflates per-image Pearson.}
Write each measured pattern as $\mathbf v_i=\bar{\mathbf m}+\mathbf r_i$, where $\bar{\mathbf m}$
is the static per-voxel mean (shared by every image and reflecting baseline and anatomical
offsets) and $\mathbf r_i$ is the stimulus-driven residual. Under the mean-retained regime
$\bar{\mathbf m}$ is present in both the target and any reasonable prediction, and because it is
identical across images it is trivially reproduced. The across-voxel correlation is then dominated
by agreement on $\bar{\mathbf m}$ rather than on the stimulus-specific $\mathbf r_i$, inflating
per-image Pearson. Z-scoring removes $\bar{\mathbf m}$, leaving $\mathbf r_i$ alone, so the metric
measures stimulus-driven fidelity.

\begin{figure}[tb]
  \centering
  \includegraphics[width=0.8\linewidth]{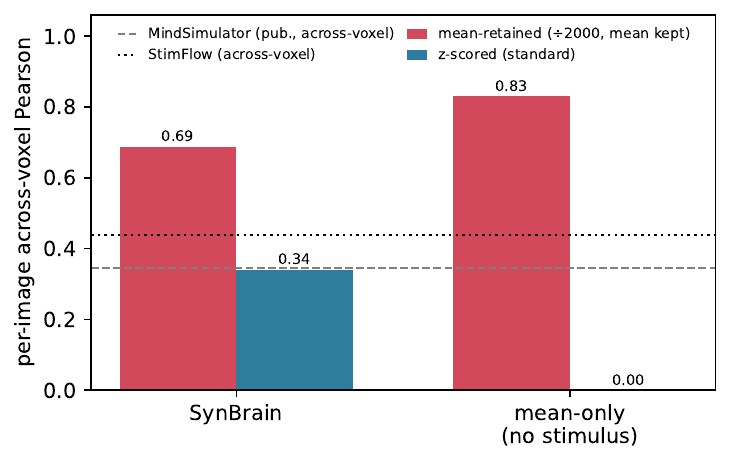}
  \caption{\textbf{Per-image, across-voxel Pearson under two normalization regimes.}
    Per-image, across-voxel Pearson (SynBrain's reported metric) for SynBrain and for a stimulus-agnostic mean-only predictor,
    under the mean-retained regime (constant division, per-voxel mean kept) and under standard
    z-scoring. MindSimulator (published) and StimFlow are shown for reference. The mean-only
    predictor scores high when the mean is retained and collapses under z-scoring, showing that
    the inflated mean-retained Pearson reflects the static per-voxel mean rather than
    stimulus-driven fidelity. The mean-only bars are computed on subject~1 (the subject with the
    mean-retained signal on hand); SynBrain bars are from our reproduction (mean-retained $=$
    originally reported $0.687$; z-scored $=0.340$).}
  \label{fig:supp-norm}
\end{figure}

\paragraph{A mean-only control isolates the effect.}
Consider a predictor that ignores the stimulus and outputs the constant per-voxel training mean
$\bar{\mathbf m}$ for every image. Under the mean-retained regime this trivial predictor already
attains a high per-image Pearson of $0.83$ on subject~1, \emph{exceeding} SynBrain's reported
$0.687$ (\cref{fig:supp-norm}, ``mean-only''), despite carrying no stimulus information; under
z-scoring its correlation collapses to chance, because the prediction equals the train mean and
loses all across-voxel variance once the mean is removed. Since the static mean alone already
yields $0.83$ in the mean-retained regime --- itself above SynBrain's reported $0.687$ --- a score
reported in that regime reflects mostly this mean rather than stimulus-specific fidelity; removing
the mean by z-scoring strips the artifact and gives the comparable $0.340$.

\paragraph{MSE moves oppositely.}
Unlike Pearson, mean-squared error is scale-dependent: the constant division shrinks the signal's
dynamic range, so the mean-retained regime also reports a much smaller MSE ($0.139$ vs.\ $0.398$
after z-scoring). The two metrics improve together only because both follow from the same
normalization, not because the synthesized signal is closer to the target.

\paragraph{Convergent evidence for the comparable value.}
Under z-scoring, our re-run of SynBrain ($0.340$) closely matches the independently reported
MindSimulator value ($0.346$); only SynBrain's mean-retained number ($0.687$) stands apart. Two
independently developed methods converging to $\approx\!0.34$ under the standard normalization,
with the single outlier accounted for by mean retention, supports treating $0.340$ as the
comparable voxel-level score.

\paragraph{Reproduction protocol.}
The effect is reproduced as follows, using the released SynBrain checkpoint and the shared
$1{,}000$-image test set restricted to the \texttt{nsdgeneral} ROI:
(i)~obtain SynBrain's synthesized betas $\hat{\mathbf v}_i$ and the measured betas $\mathbf v_i$;
(ii)~compute per-image across-voxel Pearson under the mean-retained regime (constant division,
no mean removal) and under z-scoring (per-voxel train mean/std);
(iii)~add the mean-only control by replacing $\hat{\mathbf v}_i$ with the broadcast per-voxel
train mean;
(iv)~report the mean over images. The mean-only control in \cref{fig:supp-norm} is computed on
the measured betas of subject~1, for which the mean-retained signal is available; the analysis
script is released with our code.

\paragraph{Implication for evaluation.}
Semantic-level metrics alone can mask voxel-level error: a robust frozen decoder can recover the
correct image even from a coarsely synthesized response, so high decoded-image scores do not
guarantee faithful voxel patterns. A voxel-level Pearson computed under standard z-scoring is
therefore a necessary check, and we report both levels throughout.

\end{document}